\documentclass[conference]{IEEEtran}
\IEEEoverridecommandlockouts
\usepackage{cite}
\usepackage{amsmath,amssymb,amsfonts}
\usepackage{algorithmic}
\usepackage{graphicx}
\usepackage{textcomp}
\usepackage{xcolor}
\usepackage{hyperref}
\usepackage{url}

\usepackage{algorithm}
\usepackage{algorithmic}
\usepackage{booktabs}
\usepackage{multirow}

\def\BibTeX{{\rm B\kern-.05em{\sc i\kern-.025em b}\kern-.08em
    T\kern-.1667em\lower.7ex\hbox{E}\kern-.125emX}}
\begin{document}

\title{AD-Reasoning: Multimodal Guideline-Guided Reasoning for Alzheimer’s Disease Diagnosis}

\author{
\IEEEauthorblockN{
Qiuhui Chen\textsuperscript{1},
Yushan Deng\textsuperscript{2},
Xuancheng Yao\textsuperscript{2},
Yi Hong\textsuperscript{2,*}
}
\IEEEauthorblockA{
\textsuperscript{1}School of Information Science and Engineering, East China University of Science and Technology, Shanghai, China \\
chenqh@ecust.edu.cn
}
\IEEEauthorblockA{
\textsuperscript{2}School of Computer Science, Shanghai Jiao Tong University, Shanghai, China \\
\{dengysh23, 2212582443, yi.hong\}@sjtu.edu.cn
}
\IEEEauthorblockA{
\textsuperscript{*}Corresponding author
}
}

\maketitle

\begin{abstract}
Alzheimer’s disease (AD) diagnosis requires integrating neuroimaging with heterogeneous clinical evidence and reasoning under established criteria, yet most multimodal models remain opaque and weakly guideline-aligned. We present \textit{AD-Reasoning}, a multimodal framework that couples structural MRI with six clinical modalities and a rule-based verifier to generate structured, NIA-AA-consistent diagnoses. AD-Reasoning combines modality-specific encoders, bidirectional cross-attention fusion, and reinforcement fine-tuning with verifiable rewards that enforce output format, guideline evidence coverage, and reasoning--decision consistency. We also release \textit{AD-MultiSense}, a 10,378-visit multimodal QA dataset with guideline-validated rationales built from ADNI/AIBL. On AD-MultiSense, AD-Reasoning achieves state-of-the-art diagnostic accuracy and produces structured rationales that improve transparency over recent baselines, while providing transparent rationales.
\end{abstract}


\section{Introduction}

Recent progress in AI has accelerated research on neurodegenerative diseases and computer-aided diagnosis~\cite{rajpurkar2022ai,park2023methods}. 
Accurate Alzheimer’s disease (AD) assessment, however, inherently requires \emph{multimodal} evidence rather than a single signal. 
Beyond structural MRI (sMRI) patterns such as hippocampal atrophy~\cite{frisoni2010clinical,jang2022m3t}, clinicians rely on cognitive tests (e.g., MMSE), demographic and medical history, genetic risk (e.g., APOE-$\epsilon$4), laboratory findings, and CSF biomarkers (A$\beta$, tTau, pTau) to characterize disease stage and rule out confounders~\cite{lautner2014apolipoprotein}. 
This heterogeneity motivates multimodal integration for holistic and reliable AD diagnosis~\cite{venugopalan2021multimodal}.

Prior multimodal AD models improve performance by fusing imaging with clinical measurements~\cite{chen2024alifuse}, but they often remain \emph{black boxes}, producing labels/scores with limited justification and weak alignment to clinical standards. 
Meanwhile, multimodal large language models (MLLMs) have shown strong cross-modal representation alignment and generative reasoning abilities~\cite{openai2023gpt4}. 
In medicine, existing MLLM efforts largely target unimodal or shallow multimodal tasks (e.g., imaging description/reporting)~\cite{bai2024m3d}, and they still struggle to generate disease-level, guideline-consistent diagnostic narratives that synthesize heterogeneous evidence for AD and its differential diagnosis.

A key challenge is not only fusing modalities but also ensuring the generated reasoning is \emph{verifiable} and \emph{clinically consistent}. 
Recent reinforcement fine-tuning methods such as Group Relative Policy Optimization (GRPO) stabilize optimization by comparing candidate responses within groups, showing advantages over PPO in text and vision-language settings~\cite{hu2025reinforce++,li2025optimizing}. 
However, applying GRPO to AD diagnosis is non-trivial: generic similarity rewards do not capture diagnostic validity, multimodal grounding is often under-explored, and guideline-level verification (e.g., NIA-AA) is typically absent.

To address these gaps, we propose \textbf{AD-Reasoning}, a guideline-guided multimodal reasoning framework for AD diagnosis. 
Given a patient’s sMRI and six categories of clinical data (demographics, history, cognitive assessments, laboratory tests, genetics, and CSF biomarkers), AD-Reasoning generates structured diagnostic narratives grounded in multimodal evidence. 
We design modality-specific encoders and projectors, and a bidirectional cross-attention based Multimodal Fusion Layer to explicitly model neuro-clinical interactions. 
To promote transparency and clinical compliance, we further introduce a domain-specific reinforcement fine-tuning stage under RL with verifiable rewards, where GRPO is guided by NIA-AA-aligned criteria and reasoning-consistency constraints.

Our contributions are summarized as follows:
\begin{itemize}
    \item \textbf{AD-MultiSense}: an AD-specific multimodal QA dataset combining sMRI with six clinical modalities, totaling 10,378 samples from 2,619 subjects, with guideline-validated diagnostic rationales.
    \item \textbf{AD-Reasoning}: a unified MLLM framework with modality-harmonized encoding and cross-modal fusion, plus GRPO-based reinforcement fine-tuning using NIA-AA verifiable rewards for clinically consistent generation.
    \item \textbf{Strong performance}: state-of-the-art diagnostic accuracy and improved interpretability on AD-MultiSense compared with recent multimodal baselines.
\end{itemize}

\section{Methodology}

\subsection{AD-MultiSense Dataset}
\textbf{Multimodal Data Collection}
To enable MLLMs to perform both physiological understanding and diagnostic reasoning over heterogeneous medical data, we construct a multimodal dataset that conforms to established clinical logic. Raw data are collected from the ADNI~\cite{petersen2010alzheimer} and AIBL~\cite{ellis2009australian} cohorts, covering a wide spectrum of patient characteristics and disease stages. For each subject, we acquire sMRI scans alongside six types of clinical data encompassing demographic, cognitive, and biochemical information. After aligning data across modalities and visit timepoints, we curate a total of 10,378 multimodal samples from 2,619 unique subjects. Each sample reflects a consistent physiological state at a specific visit, enabling clinically valid reasoning over disease progression.

To enhance clinical interpretability, quantitative measurements are systematically converted into standardized textual reports. For sMRI analysis, we calculate age-adjusted $z$-scores for structural volumes (e.g., hippocampal/ventricular) using population norms, with textual descriptors generated based on established thresholds: bilateral hippocampal atrophy is reported as ``mild'' ($1 \leq |z| < 1.5$), ``moderate'' ($1.5 \leq |z| < 2$), ``significant'' ($2 \leq |z| < 3$) or ``profound'' ($|z| \geq 3$). Similarly, laboratory data undergo $z$-score normalization against age/sex-matched cohorts, though only clinically significant abnormalities ($|z| > 2.0$) are included in final reports. 
Biomarkers are consistently interpreted with contextual information, and each value is accompanied by reference-based interpretation, e.g., "Amyloid beta: 858.30 pg/mL (normal)."
This quantitative-to-textual transformation bridges raw biomarker measurements with clinically meaningful narratives, enabling natural language reasoning about pathological changes while preserving data fidelity. 

\begin{figure*}[t]
    \centering
    \includegraphics[width=\linewidth]{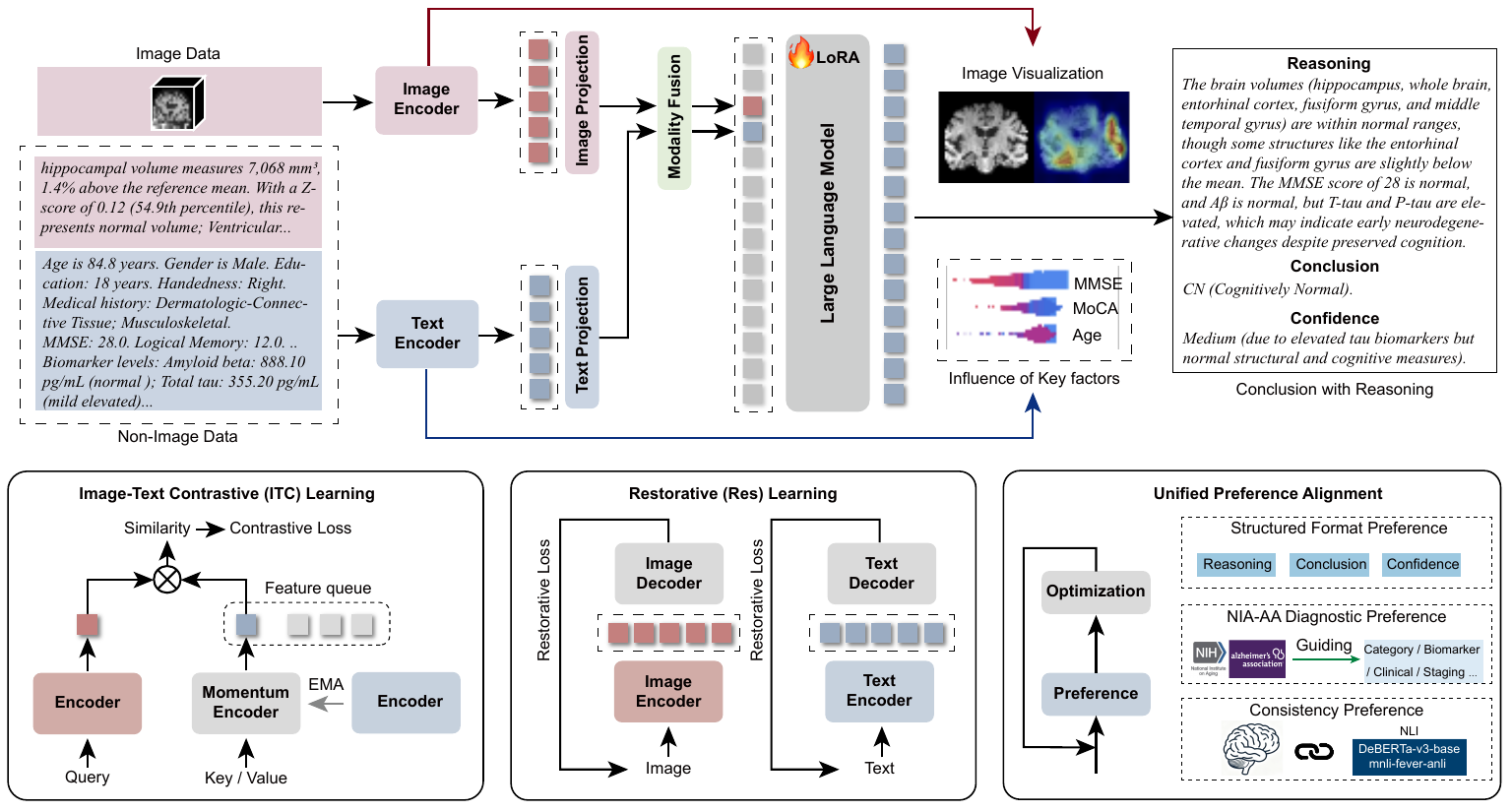}
    \caption{AD-Reasoning framework. Pretraining aligns sMRI and clinical data representations via encoders, SFT tunes LLMs using diagnostic rationales and RFT optimizes with GRPO for NIA-AA compliant structured outputs.}
    \label{fig:overview}
\end{figure*}

\begin{figure}[t]
    \centering
    \includegraphics[width=\linewidth]{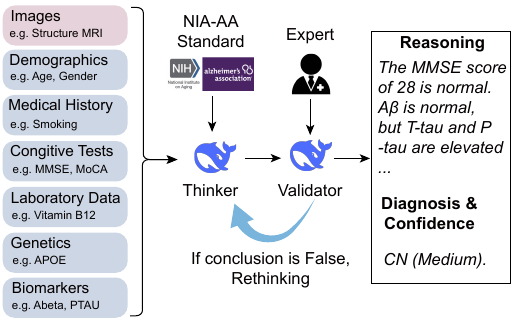}
    \caption{
    Our AD-MultiSense dataset Construction pipeline: disease-level reports are generated via evidence-augmented reasoning under clinical guidelines with self-refinement for diagnostic validity.
    }
    \label{fig:dataset}
\end{figure}

\textbf{Reasoning Generation}
Based on these raw data, we construct multimodal QA pairs from disease-level diagnostic reasoning, with the entire process shown in Fig.~\ref{fig:dataset}. The process begins by querying the \textit{Thinker} model (DeepSeek-V3) with a structured diagnostic prompt that enforces a fixed output schema consisting of \texttt{Reasoning}, \texttt{Diagnosis} (CN/MCI/Dementia), and \texttt{Confidence} (High/Medium/Low).
This is an initial response $\langle R_0, C_0 \rangle = \text{Thinker}(M, P_d)$, where $R_0$ denotes the reasoning chain, $C_0$ is the preliminary diagnosis, $M$ represents multimodal inputs (i.e., sMRIs and clinical data), and $P_d$ is the diagnosis prompt.

The \textit{Validator} module evaluates $C_0$ against ground truth diagnoses. When mismatches occur, the system triggers rethinking cycles: the Thinker regenerates reasoning using refinement prompts ($P_r$) constructed from explicit NIA-AA criteria dictionaries. These dictionaries map clinical findings to diagnostic rules, enabling targeted feedback. This iterative process continues for up to N cycles (i.e., 2), with random expert sampling providing quality control.

For cases where diagnosis remains incorrect after N iterations, the prompts with correct diagnosis ($P_c$) is explicitly provided to the Thinker, instructing it to correct its reasoning and conclusion accordingly. The Thinker then produces final reasoning $R^F$ and diagnosis $C^F$, formatted into training pairs $\langle M \circ P_d, R^F \circ C^F \rangle$ for supervised fine-tuning.

\subsection{AD-Reasoning Framework}

\subsubsection{Model Architecture}
The proposed AD-Reasoning framework primarily consists of modality-specific encoders and projectors, a Multimodal Fusion Layer (MFL), and a Large Language Model (LLM), with its overall architecture illustrated in Fig.~\ref{fig:overview}. 
Given the raw data of structural MRI scans $\mathbf{X_V} \in \mathbb{R}^{1 \times D \times H \times W}$ and clinical text data $\mathbf{X_T} \in \mathbb{R}^{L}$, they are first processed by their respective modality-specific encoders for feature extraction. 
The encoded features are then fed into modality-specific projectors to transform them into a shared dimension $d$ for alignment and compatibility with the textual embedding space of the LLM. 
This process facilitates seamless integration between multimodal features and textual tokens, formulated as:
\begin{equation}
     \mathbf{V}_{\text{sMRI}} = g_V(f_V(\mathbf{X_V})) \in \mathbb{R}^{d}, \quad
     \mathbf{T}_{\text{Clinical}} = g_T(f_T(\mathbf{X_T})) \in \mathbb{R}^{d},
\end{equation}
where $\mathbf{V}_{\text{sMRI}}$ denotes projected visual features from structural MRI, $\mathbf{T}_{\text{Clinical}}$ denotes projected clinical text features. $f_V, f_T$ denotes modality-specific encoders (image and text) and $g_V, g_T$ denotes modality-specific projectors.

\subsubsection{Multimodal Fusion Layer (MFL)}
To enable comprehensive interaction between neuroimaging and clinical modalities, we introduce an MFL comprising a Bidirectional Cross-Attention (BCA) mechanism. The projected features $\mathbf{V}_{\text{sMRI}}$ and $\mathbf{T}_{\text{Clinical}}$ are first processed by the BCA mechanism, where each modality alternately serves as Query and Key/Value to compute cross-attention:
\begin{align}
\mathbf{A}_{V \rightarrow T} &= \text{Attention}(\mathbf{T}_{\text{Clinical}}, \mathbf{V}_{\text{sMRI}}, \mathbf{V}_{\text{sMRI}}), \\
\mathbf{A}_{T \rightarrow V} &= \text{Attention}(\mathbf{V}_{\text{sMRI}}, \mathbf{T}_{\text{Clinical}}, \mathbf{T}_{\text{Clinical}}).
\end{align}

This bidirectional attention captures complex neuro-clinical dependencies, allowing visual features to inform clinical interpretation and vice versa. The attention outputs are combined with residual connections to preserve modality-specific information:
\begin{equation}
     \mathbf{T}_V = \mathbf{V}_{\text{sMRI}} + \mathbf{A}_{T \rightarrow V}, \quad
     \mathbf{T}_T = \mathbf{T}_{\text{Clinical}} + \mathbf{A}_{V \rightarrow T}.
\end{equation}

\subsubsection{Large Language Model Integration}
The final multimodal features $\mathbf{T}_V$ and $\mathbf{T}_T$ replace the placeholders \texttt{<sMRI>} and \texttt{<clinical>} in the input prompt templates. An example prompt for AD diagnosis is: 
\begin{quote}
\small
``Given the structural MRI \texttt{<sMRI>} and clinical profile \texttt{<clinical>}, what is the most probable diagnosis and supporting evidence?''
\end{quote}

The resulting input sequence $\mathbf{T}_{\text{input}} = \{\mathbf{T_Q}, \mathbf{T}_V,\mathbf{T}_T, $ $ \mathbf{T_A}\}$ is fed into the LLM, where $\mathbf{T_Q}$ denotes tokenized question derived from diagnostic templates and $\mathbf{T_A}$ denotes target answer tokens from AD diagnostic QA datasets.

The LLM parameters remain frozen during training, with only LoRA adapters updated to specialize the model for AD reasoning tasks.

\subsubsection{Training Strategy}
We employ a three-stage training strategy for AD-Reasoning, which includes Pre-training (PT), Supervised Fine-Tuning (SFT), and Reinforcement Fine-Tuning (RFT), to progressively enhance its ability
to perceive the physiological representations of each modality and integrate multimodal information for interpretable Alzheimer's disease reasoning and diagnosis.

\textbf{Pre-training (PT).}
To align sMRI and clinical representations, we pre-train the image and text encoders while keeping the projectors and the LLM inactive. We optimize an image--text contrastive (ITC) loss~\cite{radford2021learning} and adopt EMA-updated momentum encoders with a feature queue for stable negatives, following BLIP/ALBEF~\cite{li2022blip,li2021align} (momentum coefficient $m_c=0.995$). To preserve fine-grained information, we further introduce a restorative reconstruction branch that reconstructs the input image (MSE) and clinical text (token-level cross-entropy). The overall pre-training objective is:
\begin{equation}
\mathcal{L}_{\text{PT}} = \mathcal{L}_{\text{itc}} + \lambda_{\text{res}} \left( \mathcal{L}_{res}^I + \mathcal{L}_{res}^T \right)
\end{equation}
where $\mathcal{L}_{\text{itc}}$ aligns image/text features, $\mathcal{L}_{res}^I$ and $\mathcal{L}_{res}^T$ are image/text reconstruction losses, and $\lambda_{\text{res}}$ balances the restoration objective.

\textbf{Supervised Fine-Tuning (SFT).} Building upon the aligned feature representations, we conduct SFT using diagnostic QA pairs for AD reasoning. During this stage, image and text encoders are frozen, and the projection layers and LLM LoRA modules are trainable.
The optimization objective maximizes response generation likelihood:
\begin{equation}
\begin{aligned}
\mathcal{L}_{\text{SFT}} =  & -\mathbb{E}_{(\mathbf{T_Q}, \mathbf{V}_{\text{sMRI}}, \mathbf{T}_{\text{Clinical}}, \mathbf{T_A}) \sim \mathcal{D}} \\
 & \sum_{t=1}^{T} \log \pi_{\theta} \big( y_t \mid \mathbf{T_Q}, \mathbf{V}_{\text{sMRI}}, \mathbf{T}_{\text{Clinical}}, y_{<t} \big),
\end{aligned}
\end{equation}
where $\pi_{\theta}(y_t | \cdot)$ denotes the conditional probability of generating the $t$-th token $y_t$, given the prompt tokens $\mathbf{T_Q}$, modality features ($\mathbf{V}_{\text{sMRI}}$ and $\mathbf{T}_{\text{Clinical}}$), and the previously generated tokens $y_{<t}$. $\mathbf{V}_{\text{sMRI}}$ denotes visual features from structural MRI and $\mathbf{T}_{\text{Clinical}}$ encompasses all clinical texts. $\mathbf{T_Q}$ denotes question tokens and $\mathbf{T_A}$ denotes answer tokens.

\begin{table*}[t]
\scriptsize
\centering
\caption{Comparison of AD-Reasoning and baselines in terms of reasoning and diagnostic performance for Alzheimer’s disease.}
\begin{tabular}{clcccccccc}
\toprule
& \textbf{Method} & \textbf{BLEU} & \textbf{METEOR} & \textbf{ROUGE} & \textbf{BERT} & \textbf{ACC} (\%) & \textbf{AUC} (\%) & \textbf{SEN} (\%) & \textbf{SPE} (\%) \\
\midrule
\multirow{7}{*}{CN vs. CI} &
LLaVA-1.5-7B & 0.0112 & 0.1456 & 0.1023 & 0.7924 & 73.85 & 68.92 & 60.14 & 80.37 \\
& LLaVA-Med & 0.0144 & 0.1618 & 0.1168 & 0.8016 & 76.21 & 71.43 & 62.75 & 83.42 \\
& Med-PaLM-M & 0.0218 & 0.2031 & 0.1331 & 0.8181 & 79.92 & 75.76 & 66.63 & 85.85 \\
& M3d-LaMed & 0.0341 & 0.1756 & 0.1435 & 0.8128 & 82.37 & 78.95 & 69.84 & 86.21 \\
& AD-Reasoning w/o PT & 0.1873 & 0.2792 & 0.2424 & 0.8636 & 87.25 & 83.12 & 71.28 & 91.37 \\
& AD-Reasoning w/o RFT & 0.2015 & 0.2982 & 0.2617 & 0.8725 & 90.46 & 87.63 & 80.75 & 94.28 \\
& AD-Reasoning (ours) & \textbf{0.2183} & \textbf{0.3212} & \textbf{0.2851} & \textbf{0.8926} & \textbf{93.33} & \textbf{91.83} & \textbf{88.67} & \textbf{95.00} \\
\midrule
\multirow{7}{*}{CN vs. MCI} &
LLaVA-1.5-7B & 0.0108 & 0.1387 & 0.0984 & 0.7821 & 70.15 & 65.28 & 61.42 & 74.85 \\
& LLaVA-Med & 0.0138 & 0.1518 & 0.1068 & 0.7916 & 72.24 & 68.76 & 65.57 & 77.36 \\
& Med-PaLM-M & 0.0208 & 0.1931 & 0.1231 & 0.8081 & 75.13 & 72.14 & 68.41 & 80.25 \\
& M3d-LaMed & 0.0331 & 0.1656 & 0.1335 & 0.8028 & 78.02 & 74.97 & 70.79 & 81.64 \\
& AD-Reasoning w/o PT & 0.1824 & 0.2717 & 0.2369 & 0.8570 & 88.37 & 84.96 & 84.92 & 87.41 \\
& AD-Reasoning w/o RFT & 0.1961 & 0.2893 & 0.2544 & 0.8667 & 91.28 & 89.07 & 88.45 & 90.33 \\
& AD-Reasoning (ours) & \textbf{0.2123} & \textbf{0.3125} & \textbf{0.2783} & \textbf{0.8852} & \textbf{92.82} & \textbf{90.09} & \textbf{88.60} & \textbf{93.50} \\
\bottomrule
\end{tabular}
\label{tab:final_results}
\end{table*}

\begin{table}[t]
\centering
\begin{minipage}{0.5\textwidth}
\centering
\tiny
\caption{Diagnostic performance (\%) comparison between our AD-Reasoning and classification approaches for Alzheimer's disease. (Best in bold)}
\begin{tabular}{lcccccccc}
\toprule
& \multicolumn{4}{c}{\textbf{CN vs. CI}} & \multicolumn{4}{c}{\textbf{CN vs. MCI}} \\
\cmidrule(lr){2-5} \cmidrule(lr){6-9}
\textbf{Method} & \textbf{ACC} & \textbf{AUC} & \textbf{SEN} & \textbf{SPE} & \textbf{ACC} & \textbf{AUC} & \textbf{SEN} & \textbf{SPE} \\
\midrule
BERT & 84.31 & 79.42 & 85.87 & 86.48 & 82.55 & 77.35 & 82.67 & 84.42 \\
RoBerta & 86.89 & 84.41 & 82.97 & 85.03 & 85.63 & 81.42 & 80.93 & 83.84 \\
Longformer & 87.92 & 85.76 & 80.49 & 82.27 & 85.24 & 84.71 & 78.42 & 79.37 \\
IRENE & 86.03 & 77.95 & 89.14 & 65.82 & 84.18 & 75.25 & 87.35 & 63.27 \\
AD-Trans & 87.67 & 75.89 & 65.91 & 85.47 & 85.61 & 73.79 & 63.67 & 84.32 \\
Alifuse & 87.23 & 79.51 & \textbf{90.71} & 73.67 & 85.98 & 76.57 & \textbf{88.92} & 70.39 \\
Ours & \textbf{93.33} & \textbf{91.83} & 88.67 & \textbf{95.00} & \textbf{92.82} & \textbf{90.09} & 88.60 & \textbf{93.50}\\
\bottomrule
\end{tabular}
\label{tab:cls}
\end{minipage}
\hfill
\begin{minipage}{0.5\textwidth}
\centering
\scriptsize
\caption{Ablation results (\%) on the test set.}
\begin{tabular}{lllll}
\toprule
    \textbf{Task} & \textbf{ACC} & \textbf{AUC} & \textbf{SEN} & \textbf{SPE} \\
\cmidrule(r){1-5}
    \textbf{(a) Loss Terms} \\
\cmidrule(r){1-5}
    $\mathcal{L}_{\text{itc}}$ & 89.23 & 84.87 &  95.12 & 79.84  \\      $\mathcal{L}_{\text{itc}}+\mathcal{L}_{\text{res}}^I + \mathcal{L}_{\text{res}}^T$ & \textbf{93.33} & \textbf{91.83} & \textbf{88.67} & \textbf{95.00}  \\
\cmidrule(r){1-5}
    \textbf{(b) Feature Terms} \\
\cmidrule(r){1-5}
    Image & 71.24 & 54.76 & 95.33 & 12.31 \\
    Clinical & 88.83 & 82.69 & 96.91 & 67.42 \\
    Image + Clinical & \textbf{93.33} & \textbf{91.83} & \textbf{88.67} & \textbf{95.00}\\
\cmidrule(r){1-5}
    \textbf{(c) Guideline Terms} \\
\cmidrule(r){1-5}
    IWG-2 & 92.93 & 90.58 & \textbf{90.12} & 87.33 \\
    NIA-AA & \textbf{93.33} & \textbf{91.83} & 88.67 & \textbf{95.00}\\
\bottomrule
\end{tabular}
\label{tab:ablation}
\end{minipage}
\end{table}

\textbf{Reinforcement Fine-Tuning (RFT).}
To unlock the potential of the constructed dataset and enhance diagnostic reasoning capabilities, we perform Reinforcement Fine-Tuning (RFT) using Group Relative Policy Optimization (GRPO) under the RL with Verifiable Rewards (RLVR) framework. The trainable components remain consistent with the SFT stage, with the optimization objective:
\begin{equation}
\begin{aligned}
 &\max_{\pi_{\theta}} \mathbb{E}_{\mathbf{A} \sim \pi_{\theta}(\mathbf{Q})} \left[ R_{\text{RLVR}}(\mathbf{Q}, \mathbf{A}) \right]  \\
& = \bigg[ R(\mathbf{Q}, \mathbf{A}) - \beta \, \text{KL} \left[ \pi_{\theta}(\mathbf{A} \mid \mathbf{Q}) \| \pi_{\text{ref}}(\mathbf{A} \mid \mathbf{Q}) \right] \bigg]
\end{aligned}
\end{equation}
where $\pi_{\theta}$ is the policy and $\pi_{\text{ref}}$ is the SFT-tuned reference. $R$ denotes the verifiable reward function, while the KL divergence term penalizes deviation from clinically validated responses, with $\beta$ controlling the regularization strength.

For AD diagnosis where responses exhibit high clinical specificity, GRPO directly compares responses within candidate groups $\{o_1, \dots, o_G\}$. Reward normalization uses:
$\tilde{r}_i = \frac{r_i - \mu_r}{\sigma_r + \epsilon}$,
where $\mu_r$ and $\sigma_r$ are group reward statistics. This prioritizes clinically coherent responses without requiring separate critic models.

The composite reward $R = R_F + R_{\text{NIA-AA}} + R_{\text{consistency}}$  ensures diagnostic accuracy and structural consistency:

\noindent
\textit{1) Structured Format Reward ($R_F$):} Enforces compliance with AD diagnostic templates:
  \begin{verbatim}
  Reasoning: [analysis]
  Diagnosis: [CN/MCI/Dementia] 
  Confidence: [High/Medium/Low]
  \end{verbatim}
$R_F = 1.0$ only when all three tags are present and \texttt{Confidence} contains valid value.

\noindent
\textit{2) NIA-AA Diagnostic Reward ($R_{\text{NIA-AA}}$):} Provides comprehensive clinical assessment through a multi-dimensional scoring framework that evaluates diagnostic accuracy against established NIA-AA standards. The reward integrates three core components:
\begin{equation}
R_{\text{NIA-AA}} = 0.4 \cdot R_{\text{cat}} + 0.3 \cdot R_{\text{bio}} + 0.3 \cdot R_{\text{feat}} .
\end{equation}

\textbf{Diagnostic Category Alignment ($R_{\text{cat}}$)} ensures precise classification into standardized diagnostic categories (CN, MCI, Dementia) through keyword matching and exclusion criteria validation. This component evaluates both the presence of appropriate diagnostic terminology and the absence of contradictory indicators.

\textbf{Biomarker Consistency Assessment ($R_{\text{bio}}$)} quantifies the coverage and contextual accuracy of essential AD biomarkers (A$\beta$, tTau, pTau). The scoring incorporates both mention frequency and pathological status characterization (normal/abnormal patterns) based on established clinical thresholds.

\textbf{Clinical Feature Comprehensiveness ($R_{\text{feat}}$)} evaluates the depth of cognitive domain analysis across memory, executive function, visuospatial abilities, and language domains. The scoring rewards not only feature inclusion but also detailed characterization within specific subdomains.

This structured approach ensures rigorous adherence to NIA-AA diagnostic protocols while maintaining computational efficiency through weighted component integration. 

\noindent
\textit{3) Reasoning Consistency Reward ($R_{\text{consistency}}$):} To ensure logical coherence between diagnostic reasoning and final conclusions, we introduce a reasoning consistency reward in the overall reward structure. This component evaluates the alignment between the analysis in the Reasoning section and the diagnostic conclusion:

\begin{equation}
R_{\text{consistency}} = \text{NLI}(\text{Reasoning} \rightarrow \text{Diagnosis})
\end{equation}

where NLI denotes Natural Language Inference~\cite{he2021debertav3}, implemented using a pre-trained entailment model that scores the degree to which the reasoning text supports the diagnostic conclusion. The reward $R_{\text{consistency}} \in \{0, 0.5, 1.0\}$ corresponds to contradiction, neutral/weak entailment, and strong entailment, respectively. This prevents logical inconsistencies where, for example, the reasoning describes normal biomarker profiles but concludes with "Dementia," ensuring that diagnostic conclusions are well-supported by the preceding clinical analysis.

This enhanced reward structure ensures comprehensive alignment with NIA-AA diagnostic standards while maintaining computational efficiency and logical coherence. The format reward $R_F$ guarantees structural integrity, $R_{\text{NIA-AA}}$ evaluates clinical content validity, and $R_{\text{consistency}}$ ensures logical alignment between analysis and conclusions.

\vspace{-2mm}

\section{Experiments}
\vspace{-2mm}
We conduct all experiments on a server equipped with four NVIDIA RTX 3090 24GB GPUs. For the LLM, we choose LLaMA 3.2-1B~\cite{grattafiori2024llama3herdmodels} and integrate the LoRA modules~\cite{hu2022lora} with a rank of 8 for fine-tuning.
For the visual modality, a 3D Vision Transformer~\cite{dosovitskiy2020image} is used with input size $128 \times 128 \times 128$ and patch size $16 \times 16 \times 16$. 
For the textual modality, we use a Longformer Transformer~\cite{beltagy2020longformer}.  
The PT, SFT and RFT stages are each trained for 100 epochs, while the RFT stage is trained using the open-source Trainer framework.

The effectiveness of multimodal reasoning and diagnosis is evaluated from two sides. 1) The descriptive accuracy of the generated diagnostic text is assessed using natural language generation (NLG) metrics, including BLEU, METEOR, ROUGE, and BERT. 
2) The classification accuracy of Alzheimer disease categories in the responses is evaluated using diagnosis accuracy (ACC), Area
Under Curve (AUC), sensitivity (SEN), and specificity (SPE).

Following established clinical guidelines~\cite{mckhann2011diagnosis,jack2018nia}, we evaluate our model on two classification tasks. The first task distinguishes cognitively normal (CN) individuals from those with cognitive impairment (CI), including both mild cognitive impairment (MCI) and Alzheimer’s disease (AD). The second one focuses on differentiating CN from MCI, which is a critical stage for the early identification of AD. We split the dataset {\it subject-wise} into training, validation, and test sets with proportions of 70\%, 10\%, and 20\%, respectively. 
All structural MRI scans underwent standardized preprocessing, including skull stripping to remove non-brain tissues and intensity normalization to harmonize voxel value distributions across scanners. 

\begin{figure}[t]
    \centering
    \includegraphics[width=\linewidth]{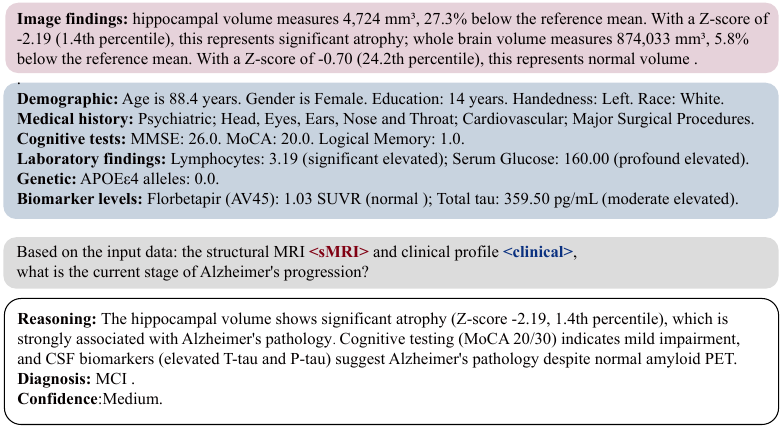}
    \caption{Inference example of AD-reasoning.}
    \label{fig:example}
\end{figure}

\subsection{Quantitative Analysis}
Given the absence of multimodal models specialized for AD integrating neuroimaging and comprehensive clinical data, we adapt comparative frameworks by measuring sMRI volumes and generating descriptions, representing clinical profiles as structured text narratives. 
Table~\ref{tab:final_results} benchmarks AD-Reasoning against four state-of-the-art MLLMs: LLaVA-1.5-7B~\cite{liu2023visual}, LLaVA-Med~\cite{li2023llava}, Med-PaLM-M~\cite{tu2024towards} and M3D-LaMed~\cite{bai2024m3d}. 
These models represent the current frontier in medical multimodal reasoning. 

As shown in Table~\ref{tab:cls}, to evaluate the performance of our model, we select three prominent text-only baselines(e.g., BERT~\cite{devlin2018bert}, Roberta~\cite{liu2019roberta}, and Longformer~\cite{beltagy2020longformer}) and three recent transformer-based models that fuse multimodal information for classification(e.g.,  IRENE~\cite{zhou2023transformer}, AD-Trans~\cite{yu2024transformer}, and Alifuse~\cite{chen2024alifuse}).

The results demonstrate that AD-Reasoning outperforms these leading models, excelling not only in natural language generation but also in clinical evaluation. This indicates the superior capability of AD-Reasoning in both descriptive and diagnostic reasoning tasks in multimodal scenarios.
Furthermore, Table~\ref{tab:final_results} also presents ablation studies to investigate the impact of physiological-level pre-training and RFT-based post-training on the model’s performance. 
The results show that removing either component leads to a noticeable decline in performance. 
Specifically, the findings highlight two key insights: 
1) Pre-training enables the model to extract and align high-quality, modality-specific representations while preserving fine-grained information through restoration loss, establishing a robust foundation for cross-modal reasoning.
2) The RFT stage based on GRPO further unleashes the potential of the constructed data and enhances the model’s AD diagnostic diagnostic performance, enabling deeper and more effective cross-modal reasoning.

\subsection{Qualitative Analysis and Ablation Study}

AD-Reasoning demonstrates a robust ability to integrate and analyze data from multiple modalities to arrive at comprehensive diagnoses. This integration allows for mutual corroboration among the modalities, enhancing diagnostic accuracy, as shown in Fig.~\ref{fig:example}. 
AD-Reasoning effectively synthesizes information from sMRI and clinical non-image data to diagnose alzheimer's disease conditions. Each modality provides unique insights that collectively strengthen the diagnostic conclusion. 
The model frequently employs terms, e.g.,``indicates" and ``associated with", highlighting its capability to identify and utilize evidence from each modality to substantiate the final diagnosis. This approach demonstrates AD-Reasoning’s proficiency in extracting relevant features from each dataset, ensuring that the diagnostic reasoning is well-founded and comprehensive.

The ablation studies in Table~\ref{tab:ablation} demonstrate the effectiveness of both contrastive and restorative learning modules in the pre-training, as well as the necessity of
complete modality integration. The integration of $\mathcal{L}_{\text{itc}}$and $\mathcal{L}_{\text{res}}$ significantly enhances the results, validating our initial intention to design these mechanisms to facilitate modality fusion and adjust the contribution levels of different modalities across AD-stage diagnosis. The presence of all modalities results in the best performance. Removing any single modality leads to reduced scores. This underscores the importance of multimodal integration for optimal outcomes. 
Both guideline variants are competitive; NIA-AA improves overall ACC/AUC and specificity in our setting, while IWG-2 yields slightly higher sensitivity.

\section{Conclusion}
We presented AD-Reasoning, a guideline-guided multimodal framework that integrates sMRI with diverse clinical information to generate structured and interpretable AD diagnoses. We also introduced AD-MultiSense, which provides guideline-validated diagnostic rationales via quantitative-to-text transformation and NIA-AA-guided refinement. Our multimodal fusion design and GRPO-based reinforcement fine-tuning with verifiable rewards explicitly encourage NIA-AA compliance and reasoning--decision consistency. Experiments on AD-MultiSense demonstrate state-of-the-art diagnostic performance and improved interpretability compared with recent multimodal baselines. These results suggest a practical path toward clinically grounded and explainable decision support for early AD screening and longitudinal monitoring.

\bibliographystyle{IEEEbib}
\bibliography{icme2026references}

\begin{thebibliography}{10}

\bibitem{rajpurkar2022ai}
Pranav Rajpurkar, Emma Chen, Oishi Banerjee, and Eric~J Topol,
\newblock ``Ai in health and medicine,''
\newblock {\em Nature medicine}, vol. 28, no. 1, pp. 31--38, 2022.

\bibitem{park2023methods}
Seong~Ho Park, Kyunghwa Han, Hye~Young Jang, Ji~Eun Park, June-Goo Lee, Dong~Wook Kim, and Jaesoon Choi,
\newblock ``Methods for clinical evaluation of artificial intelligence algorithms for medical diagnosis,''
\newblock {\em Radiology}, vol. 306, no. 1, pp. 20--31, 2023.

\bibitem{frisoni2010clinical}
Giovanni~B Frisoni, Nick~C Fox, Clifford~R Jack~Jr, Philip Scheltens, and Paul~M Thompson,
\newblock ``The clinical use of structural mri in alzheimer disease,''
\newblock {\em Nature reviews neurology}, vol. 6, no. 2, pp. 67--77, 2010.

\bibitem{jang2022m3t}
Jinseong Jang and Dosik Hwang,
\newblock ``M3t: three-dimensional medical image classifier using multi-plane and multi-slice transformer,''
\newblock in {\em Proceedings of IEEE/CVF CVPR}, 2022, pp. 20718--20729.

\bibitem{lautner2014apolipoprotein}
Ronald Lautner, Sebastian Palmqvist, Niklas Mattsson, Ulf Andreasson, Anders Wallin, Erik P{\aa}lsson, Joel Jakobsson, Sanna-Kaisa Herukka, Rikard Owenius, Bob Olsson, et~al.,
\newblock ``Apolipoprotein e genotype and the diagnostic accuracy of cerebrospinal fluid biomarkers for alzheimer disease,''
\newblock {\em JAMA psychiatry}, vol. 71, no. 10, pp. 1183--1191, 2014.

\bibitem{venugopalan2021multimodal}
Janani Venugopalan, Li~Tong, Hamid~Reza Hassanzadeh, and May~D Wang,
\newblock ``Multimodal deep learning models for early detection of alzheimer’s disease stage,''
\newblock {\em Scientific reports}, vol. 11, no. 1, pp. 3254, 2021.

\bibitem{chen2024alifuse}
Qiuhui Chen, Xinyue Hu, Zirui Wang, and Yi~Hong,
\newblock ``Alifuse: Aligning and fusing multi-modal medical data for computer-aided diagnosis,''
\newblock {\em BIBM}, 2024.

\bibitem{openai2023gpt4}
OpenAI,
\newblock ``Gpt-4 technical report,'' 2023.

\bibitem{bai2024m3d}
Fan Bai, Yuxin Du, Tiejun Huang, Max Q-H Meng, and Bo~Zhao,
\newblock ``M3d: Advancing 3d medical image analysis with multi-modal large language models,''
\newblock {\em arXiv preprint arXiv:2404.00578}, 2024.

\bibitem{hu2025reinforce++}
Jian Hu,
\newblock ``Reinforce++: A simple and efficient approach for aligning large language models,''
\newblock {\em arXiv preprint arXiv:2501.03262}, 2025.

\bibitem{li2025optimizing}
Xuying Li, Zhuo Li, Yuji Kosuga, and Victor Bian,
\newblock ``Optimizing safe and aligned language generation: A multi-objective grpo approach,''
\newblock {\em arXiv preprint arXiv:2503.21819}, 2025.

\bibitem{petersen2010alzheimer}
Ronald~Carl Petersen, Paul~S Aisen, Laurel~A Beckett, Michael~C Donohue, Anthony~Collins Gamst, Danielle~J Harvey, et~al.,
\newblock ``Alzheimer's disease neuroimaging initiative (adni): clinical characterization,''
\newblock {\em Neurology}, vol. 74, no. 3, pp. 201--209, 2010.

\bibitem{ellis2009australian}
Kathryn~A Ellis, Ashley~I Bush, David Darby, et~al.,
\newblock ``The australian imaging, biomarkers and lifestyle (aibl) study of aging: methodology and baseline characteristics of 1112 individuals recruited for a longitudinal study of alzheimer's disease,''
\newblock {\em International psychogeriatrics}, vol. 21, no. 4, pp. 672--687, 2009.

\bibitem{radford2021learning}
Alec Radford, Jong~Wook Kim, Chris Hallacy, and et~al.,
\newblock ``Learning transferable visual models from natural language supervision,''
\newblock in {\em International conference on machine learning}. PMLR, 2021, pp. 8748--8763.

\bibitem{li2022blip}
Junnan Li, Dongxu Li, Caiming Xiong, and Steven Hoi,
\newblock ``Blip: Bootstrapping language-image pre-training for unified vision-language understanding and generation,''
\newblock in {\em International Conference on Machine Learning}. PMLR, 2022, pp. 12888--12900.

\bibitem{li2021align}
Junnan Li, Ramprasaath Selvaraju, Akhilesh Gotmare, Shafiq Joty, Caiming Xiong, and Steven Chu~Hong Hoi,
\newblock ``Align before fuse: Vision and language representation learning with momentum distillation,''
\newblock {\em NeurIPS}, vol. 34, pp. 9694--9705, 2021.

\bibitem{he2021debertav3}
Pengcheng He, Jianfeng Gao, and Weizhu Chen,
\newblock ``Debertav3: Improving deberta using electra-style pre-training with gradient-disentangled embedding sharing,''
\newblock {\em arXiv preprint arXiv:2111.09543}, 2021.

\bibitem{grattafiori2024llama3herdmodels}
Aaron~Grattafiori et~al.,
\newblock ``The llama 3 herd of models,'' 2024.

\bibitem{hu2022lora}
Edward~J Hu, Yelong Shen, Phillip Wallis, Zeyuan Allen-Zhu, Yuanzhi Li, Shean Wang, Lu~Wang, Weizhu Chen, et~al.,
\newblock ``Lora: Low-rank adaptation of large language models.,''
\newblock {\em ICLR}, vol. 1, no. 2, pp. 3, 2022.

\bibitem{dosovitskiy2020image}
Alexey Dosovitskiy, Lucas Beyer, Alexander Kolesnikov, Dirk Weissenborn, Xiaohua Zhai, Thomas Unterthiner, Mostafa Dehghani, et~al.,
\newblock ``An image is worth 16x16 words: Transformers for image recognition at scale,''
\newblock {\em arXiv:2010.11929}, 2020.

\bibitem{beltagy2020longformer}
Iz~Beltagy, Matthew~E Peters, and Arman Cohan,
\newblock ``Longformer: The long-document transformer,''
\newblock {\em arXiv preprint arXiv:2004.05150}, 2020.

\bibitem{mckhann2011diagnosis}
Guy~M McKhann, David~S Knopman, Howard Chertkow, Bradley~T Hyman, Clifford~R Jack~Jr, Claudia~H Kawas, William~E Klunk, Walter~J Koroshetz, Jennifer~J Manly, Richard Mayeux, et~al.,
\newblock ``The diagnosis of dementia due to alzheimer's disease: recommendations from the national institute on aging-alzheimer's association workgroups on diagnostic guidelines for alzheimer's disease,''
\newblock {\em Alzheimer's \& dementia}, vol. 7, no. 3, pp. 263--269, 2011.

\bibitem{jack2018nia}
Clifford~R Jack~Jr, David~A Bennett, Jason Blennow, et~al.,
\newblock ``Nia-aa research framework: toward a biological definition of alzheimer's disease,''
\newblock {\em Alzheimer's \& dementia}, vol. 14, no. 4, pp. 535--562, 2018.

\bibitem{liu2023visual}
Haotian Liu, Chunyuan Li, Qingyang Wu, and Yong~Jae Lee,
\newblock ``Visual instruction tuning,''
\newblock {\em Advances in neural information processing systems}, vol. 36, pp. 34892--34916, 2023.

\bibitem{li2023llava}
Chunyuan Li, Cliff Wong, Sheng Zhang, Naoto Usuyama, Haotian Liu, Jianwei Yang, Tristan Naumann, Hoifung Poon, and Jianfeng Gao,
\newblock ``Llava-med: Training a large language-and-vision assistant for biomedicine in one day,''
\newblock {\em Advances in Neural Information Processing Systems}, vol. 36, pp. 28541--28564, 2023.

\bibitem{tu2024towards}
Tao Tu, Shekoofeh Azizi, Danny Driess, Mike Schaekermann, Mohamed Amin, Pi-Chuan Chang, Andrew Carroll, Charles Lau, Ryutaro Tanno, Ira Ktena, et~al.,
\newblock ``Towards generalist biomedical ai,''
\newblock {\em Nejm Ai}, vol. 1, no. 3, pp. AIoa2300138, 2024.

\bibitem{devlin2018bert}
Jacob Devlin, Ming-Wei Chang, Kenton Lee, and Kristina Toutanova,
\newblock ``Bert: Pre-training of deep bidirectional transformers for language understanding,''
\newblock {\em arXiv:1810.04805}, 2018.

\bibitem{liu2019roberta}
Yinhan Liu,
\newblock ``Roberta: A robustly optimized bert pretraining approach,''
\newblock {\em arXiv preprint arXiv:1907.11692}, vol. 364, 2019.

\bibitem{zhou2023transformer}
Hong-Yu Zhou, Yizhou Yu, and Chengdi Wang,
\newblock ``A transformer-based representation-learning model with unified processing of multimodal input for clinical diagnostics,''
\newblock {\em Nature Biomedical Engineering}, pp. 1--13, 2023.

\bibitem{yu2024transformer}
Qi~Yu, Qian Ma, Lijuan Da, Jiahui Li, Mengying Wang, Andi Xu, Zilin Li, and Wenyuan Li,
\newblock ``A transformer-based unified multimodal framework for alzheimer's disease assessment,''
\newblock {\em Computers in Biology and Medicine}, vol. 180, pp. 108979, 2024.

\end{thebibliography}

\end{document}